\documentclass[10pt, a4paper]{article}

\usepackage[final]{lrec2026} % this is the new style
\usepackage{tabularx}
\usepackage{multirow}
\usepackage{latexsym}
\usepackage{url}
\usepackage{xurl}
\usepackage{hyperref}
\usepackage{polyglossia}
\usepackage{fontspec}
\ifluatex\usepackage{fontspec}\fi
\usepackage[normalem]{ulem}
\newfontface{\bng}{kalpurush.ttf}

\usepackage{inconsolata}
\usepackage{graphicx}
\usepackage{caption}
\usepackage{colortbl}
\usepackage{xcolor}
\usepackage{tabularray}
\usepackage{arydshln}
\usepackage{enumitem}

\title{Bangla Key2Text: Text Generation from Keywords for a Low Resource Language}

\name{Tonmoy Talukder\textsuperscript{1}, G M Shahariar\textsuperscript{2}} 

\address{
\textsuperscript{1}Ahsanullah University of Science and Technology \\
\textsuperscript{2}University of California - Riverside \\
tonmoytalukder.cs@gmail.com, gshah010@ucr.edu
}

\abstract{
This paper introduces \textit{Bangla Key2Text}, a large-scale dataset of $2.6$ million Bangla keyword--text pairs designed for keyword-driven text generation in a low-resource language. The dataset is constructed using a BERT-based keyword extraction pipeline applied to millions of Bangla news texts, transforming raw articles into structured keyword--text pairs suitable for supervised learning. To establish baseline performance on this new benchmark, we fine-tune two sequence-to-sequence models, \texttt{mT5} and \texttt{BanglaT5}, and evaluate them using multiple automatic metrics and human judgments. Experimental results show that task-specific fine-tuning substantially improves keyword-conditioned text generation in Bangla compared to zero-shot large language models. The dataset, trained models, and code are publicly released to support future research in Bangla natural language generation and keyword-to-text generation tasks.
\\ \newline
}

\begin{document}

\maketitleabstract

\section{Introduction}
Text generation is a significant and complex task within the realm of natural language generation (NLG) \citep{garbacea2020neural}. The text generation task can be characterized as generating an expected output sequence based on a given input sequence, which is also known as sequence-to-sequence (seq2seq) modeling \citep{sutskever2014sequence}. 
\begin{table}[t!]
% \centering
\vfill
\begin{tabular}{l|l} 
\cellcolor[HTML]{B0DEFE}\textbf{Keys} &  \textbf{[B]} {\bng \colorbox[HTML]{FB9393}{জমজমাট}, \colorbox[HTML]{FB9393}{মেলা}, \colorbox[HTML]{FB9393}{শুক্রবার},} \\
  & {\bng \colorbox[HTML]{FB9393}{গতকাল}} \\
 & \textbf{[E]} crowded, fair, Friday,\\
 & yesterday\\
\hdashline
\cellcolor[HTML]{B0FFA5}\textbf{Text} & \textbf{[B]} {\bng \colorbox[HTML]{F7FFA5}{গতকাল শুক্রবার মেলা}} \\  
& {\bng  \colorbox[HTML]{F7FFA5}{জমজমাট হয়েছে।}} \\
& \textbf{[E]} Yesterday was Friday and \\ 
& the fair was crowded.\\
\hline
\cellcolor[HTML]{B0DEFE}\textbf{Keys} &  \textbf{[B]} {\bng \colorbox[HTML]{FB9393}{সাধারণ}, \colorbox[HTML]{FB9393}{অভ্যস্ত}, \colorbox[HTML]{FB9393}{যাপনে},} \\
  & {\bng \colorbox[HTML]{FB9393}{যশোদাবেন}, \colorbox[HTML]{FB9393}{জীবন}} \\
 & \textbf{[E]} simple, accustomed, living, \\
 & Yashodaben, life\\
\hdashline
\cellcolor[HTML]{B0FFA5}\textbf{Text} & \textbf{[B]} {\bng \colorbox[HTML]{F7FFA5}{খুবই ধর্মপ্রাণ যশোদাবেন একেবারে} } \\  
& {\bng \colorbox[HTML]{F7FFA5}{সাধারণ জীবন যাপনে অভ্যস্ত।}} \\ 
& \textbf{[E]} Yashodaben is accustomed\\
 & to living a simple life.\\
\end{tabular}
\caption{Few examples of \colorbox[HTML]{F7FFA5}{generated Bangla texts} by our fine-tuned \textbf{mT5} model from the provided unordered \colorbox[HTML]{FB9393}{keywords}. \textbf{B} represents the keywords or texts in Bangla and \textbf{E} represents their corresponding English translation.}
\vfill
\label{table:1.1}
\end{table}
In this approach, a seq2seq model uses an encoder-decoder architecture where the encoder converts the input text (whether in the form of sequence or keywords) into a fixed-sized vector, which is then mapped to the desired target sequence by the decoder. Natural language generation models have been significantly improved due to the advancement in deep learning \citep{lecun2015deep}, enabling machines to understand and produce human-like language. The notion of seq2seq modeling has paved the way for the rapid advancement of natural language generation systems such as machine translation, summarization, question answering, and dialogue systems.

Keywords are concise representations of the content found within a document \citep{siddiqi2015keyword}. They often consist of one or more words and help to capture the essence of the text. By incorporating keywords into the text generation process, semantic coherence can be preserved, resulting in higher output quality. \textit{Keywords-to-text} generation approach is proven to be useful in several NLG tasks such as generating dialogue responses \citep{song2019generating,zhang2018personalizing}, generating method names \citep{ge2021keywords}, producing abstractive summaries \citep{li2020keywords}, and generating image captions \citep{yong2021keywords}. The use of keywords aids in the generation process, ensuring that the generated text remains relevant and meaningful.

Bangla is considered a low-resource language when it comes to language processing, though it is the seventh among the most spoken languages around the world \footnote{\href{https://www.ethnologue.com/insights/ethnologue200/}{ethnologue.com/insights/ethnologue200}}. While several studies on \textit{keywords-to-text} generation have been conducted in languages such as English \cite{ozbal2013brainsup,shen2022kwickchat,nie2022lexical,mishra2020template}, Japanese \citep{uchimoto2002text}, and Chinese \citep{song2019neural}, based on our comprehensive investigation, we have not come across any existing studies on keywords-to-text generation in Bangla. Although large language models (LLMs) have achieved notable success in general-purpose text generation, we found that particularly in Bangla, these models in their quantized form, often underperform when given unordered keyword inputs without in-context learning or fine-tuning. Our approach addresses this gap through a lightweight keyword extractor and task-specific fine-tuned smaller language models that are more effective, efficient, and reproducible. In summary, we have made the following contributions:
\begin{itemize}[left=0pt]
\item We introduce \textbf{Bangla Key2Text}, a large-scale dataset of $2.6$ million Bangla keyword-text pairs - the first and largest resource of its kind for Bangla keyword-to-text generation (Section \ref{dataset}).
\item We develop a \texttt{BERT}-based keyword extractor for identifying contextually salient Bangla keywords and fine-tune two encoder-decoder models, \texttt{mT5} and \texttt{BanglaT5}, for generating coherent, keyword-faithful Bangla text (Sections \ref{keywords} and \ref{generation}). We show a few examples in Table \ref{table:1.1}.
\item We conduct extensive evaluations, including on unseen and dialectal data, and benchmark our models against multiple 4-bit quantized open-source LLMs (LLaMA, Phi, Gemma, Mistral, Qwen, etc.), demonstrating that our fine-tuned models consistently outperform them across standard NLG metrics (Section \ref{text-gen-results}).
\item We analyze model behaviors such as cross-lingual transfer when mixing Bangla and English keywords, highlighting new opportunities for future research in multilingual keyword-driven text generation (Section \ref{text-gen-results}).
\end{itemize}

While our work includes keyword extraction and text generation models, the primary contribution of this paper is the Bangla Key2Text dataset, which provides the first large-scale benchmark for keyword-to-text generation in Bangla. The proposed models serve primarily as baseline systems to demonstrate the usability of the dataset and establish reference performance for future research.
We have made our dataset, code, and fine-tuned models publicly available\footnote{\url{https://github.com/TonmoyTalukder/Bangla-Key2Text}}.

\begin{table*}[h]
\centering
\resizebox{\textwidth}{!}
{\begin{tabular}{l|l}
\hline
\multicolumn{1}{c|}{\textbf{Keywords}} & \multicolumn{1}{c}{\textbf{Text}}               \\ 
\hline
% \textbf{[B]} {\bng জমি, কেনেন, টমেটো, গাজর, বিক্রির} & \textbf{[B]} {\bng গাজর-টমেটো বিক্রির টাকায় কেনেন ৫৪ শতক জমি।} \\ 
% \textbf{[E]} land, bought, tomatoes, carrots, selling & \textbf{[E]} He bought 54 decimal of land \\ & with the money from selling carrots and tomatoes.  \\
% \hline
\textbf{[B]} {\bng শীতের, কম, আগে, সম্ভাবনা}  &  \textbf{[B]} {\bng শীতের আগে সেই সম্ভাবনা কম।}           \\ 
\textbf{[E]} winter, slim, before, chance & \textbf{[E]} The chance is slim before winter. \\
\hline
\textbf{[B]} {\bng খবর, কোত্থেকে, বানোয়াট}  & \textbf{[B]} {\bng তারা কোত্থেকে পেল এসব বানোয়াট খবর?}     \\ 
\textbf{[E]} news, where, fake & \textbf{[E]} Where did they get these fake news? \\
\hline
\textbf{[B]} {\bng পানকৌড়ি, সাঁতার, হাওরের, জলে}         & \textbf{[B]} {\bng হাওরের শান্ত জলে আপন মনে সাঁতার কাটে পানকৌড়ি।}                    \\ 
\textbf{[E]} Cormorants, swim, lake,  water & \textbf{[E]} Cormorants swim at their own pace in the calm waters of the lake. \\
\hline
\textbf{[B]} {\bng ছাত্রীও, বক্তব্য, দেয়নি, নির্যাতিত} & \textbf{[B]}{\bng এমনকি নির্যাতিত ছাত্রীও কোনো বক্তব্য দেয়নি।}  \\
\textbf{[E]} student, statement, not, abused & \textbf{[E]} Even the abused student did not give any statement. \\
\hline
\end{tabular}}
\caption{Few keywords-text pairs from the \textbf{Bangla Key2Text} dataset.}
\label{table:4.1}
\end{table*}

\section{Related Works}
Several recent studies have explored text generation from keywords using different approaches in languages other than Bangla. \citet{uchimoto2002text} proposed a statistical tri-gram model trained on Japanese data to generate coherent keyword-based sentences. Similarly, \citet{song2019neural} proposed an attention-based encoder-decoder model trained on large Chinese dataset to generate coherent Chinese sentences from keywords. \citet{ozbal2013brainsup} introduced a rule-based approach that combines sentence fragments to generate creative sentences from keywords, using defined structural rules. \citet{shen2022kwickchat} developed an augmentative and alternative communication (AAC) system using sequence-to-sequence models with attention, which generates context-aware personalized sentences from keywords based on AAC user data. \citet{nie2022lexical} proposed neural models with attention mechanisms to better control the lexical complexity in keyword-to-text generation. Their method offers fine control over the use of rare vocabulary, leading to more readable and user-tailored sentences. \citet{mishra2020template} also presented a neural approach that generates text from keywords by following part-of-speech (POS)-based templates, with attention mechanisms ensuring alignment between structure and content. \citet{park2018lstm} applied a generative adversarial network (GAN) framework, where an LSTM-based encoder-decoder served as the generator and a Bi-LSTM with self-attention acted as the discriminator. Both components were trained adversarially to improve keyword-to-text generation.

\section{Methodology}
The proposed methodology of this work consists of three important steps: (a) dataset construction, (b) keywords extraction, and (c) text generation. We gathered $2.6$ million Bangla texts from a large public dataset, and then extracted important keywords to curate a dataset of $2.6$ million Bangla keywords-text pairs. After that we fine-tuned two pre-trained language models using $2$ million keywords-text pairs, while a separate set of $500K$ and $100K$ keywords-text pairs were used for validation and test purpose. We provide a detailed description of each step below.

\subsection{Dataset Development}\label{dataset}
To the best of our knowledge, there is currently no dataset specifically designed for the purpose of generating Bangla text from keywords. To address this gap, we developed a Bangla keywords-text pairs dataset named \textbf{Bangla Key2Text} (Bangla Keywords to Text) encompassing a total of $2.6$ million keywords-text pairs. We elaborate on the dataset development process below.\\

Although the underlying news articles are publicly available, constructing a keyword-to-text dataset at this scale requires a dedicated pipeline for keyword extraction, filtering, and dataset structuring. Our contribution lies in transforming raw Bangla text corpora into a structured dataset of keyword–text pairs suitable for supervised training and evaluation, including curated train, validation, and test splits as well as manually annotated evaluation data.\\

\noindent\textbf{(a) Data Collection.}
We utilized the \textbf{Bangla Newspaper Dataset} \citep{zabir_al_nazi_2020} that consists of $409.5$K Bangla news articles. For our study, we randomly chose $2.6$ million texts from those articles. In the 2.6 million data, the maximum number of sentences in a text is $67$ and the minimum is $1$. There are $2$ to $2.25$ sentences on average. We removed special characters and HTML tags. The Bangla newspaper corpus provides raw articles rather than structured keyword–text pairs. Therefore, we apply preprocessing and automatic keyword extraction to convert the texts into paired inputs consisting of unordered keywords and corresponding sentences suitable for supervised keyword-to-text generation.\\

\noindent \textbf{(b) Data Split.} We partitioned $2.6$ million texts into three groups: training, validation, and test. The training set comprises $2$ million texts, the validation set contains $500K$ texts, and the test set consists of $100K$ texts. In addition to that, we curated a separate evaluation dataset by manually extracting keywords from the first $10K$ texts from the $100K$ test set. The reason is twofold: (i) assess the effectiveness of our proposed keyword extraction method and (ii) determine the appropriate decoding technique for the text generation models. This manual extraction process involved two annotators, resulting in two sets of keywords for each text. We formatted each text by separating every word with a comma and presented the dataset to the annotators. They were instructed to retain only the essential keywords while discarding the less significant ones, utilizing their own discretion. Subsequently, another annotator was enlisted to choose the final set of keywords from the two sets provided by the initial annotators. 
% Further details on the annotator recruitment process can be found in Appendix \ref{app:H}.

\noindent \textbf{(c) Key2Text Dataset.} Based on the experimental results found for keyword extraction (details in section \ref{keywords}), we employed \texttt{BanglaBERT} \citep{bhattacharjee-etal-2022-banglabert} based extractor to extract keywords from all the $2.6$ million texts (train, test and validation sets) and curated a dataset consisting of Bangla keywords-text pairs (\textit{Bangla Key2Text}).

\noindent\textbf{(d) Dataset Statistics.} The maximum number of extracted keywords in a text is $259$, and the average number of extracted keywords per text is $9.381$. The maximum number of words in a text is $437$, and the average number of words per text is $15.142$. The total number of words (text length) is $40487220$ and the extracted number of keywords is $24414856$ over the 2.6 million data. The ratio of key word length and text length is $0.6030$. The top five frequently occurring keywords in the dataset are as follows - {\bng কী} (what): $272$ times, {\bng সূত্র} (source): $270$ times, {\bng প্রতিনিধি} (representative): $212$ times, {\bng এএফপি} (AFP): $196$ times, and {\bng না} (no): $191$ times. We show some instances from the dataset in Table \ref{table:4.1}.
 % and discuss potential applications of the dataset in Appendix \ref{app:G}

\subsection{Keywords Extraction}\label{keywords}
We have developed a pre-trained BERT \citep{devlin-etal-2019-bert} based keyword extractor consisting of two main steps: \textit{keywords scoring} and \textit{keywords selection}. Extracting nine keywords per text in average for $2.6$ million data manually is a very difficult task which necessitates the development of automatic extraction systems. Moreover, as the extracted keywords are contained in the text, rather than random selection, we tried to extract those that contribute most to the context of the text. Our proposed keyword extractor first assigns importance score to each word in a text, then select and extract keywords with top scores. The extracted keywords are returned in an unordered format.
\subsubsection{Experimental Settings} 
To evaluate the effectiveness of our keyword extraction method, we compare it with both neural and traditional keyword extraction baselines, including two Bangla BERT-based models and two widely used unsupervised methods (TextRank and YAKE). The BERT based models were utilized for inference using their default hyper-parameters in the keyword extraction phase. No fine-tuning was performed. The tokenized text is passed through the pre-trained model to obtain the \textit{last hidden state} as the token embeddings, which had a shape of $512\times768$. By computing the average of the embeddings across the $512$ tokens, we derived a mean embedding of shape $768$.

\subsubsection{Keywords Scoring}
To assign importance score, we utilized token embeddings using the pre-trained BERT model. We first calculated the average of all word embeddings in a text. Then for each word, we measured cosine similarity score between the word embedding and mean embedding which we considered as the importance score. We observed that the embeddings of important words are closer to the mean embedding. A high cosine similarity score between a word's embedding and the mean embedding of the text indicates that the word's embedding is pointing in a similar direction as the mean embedding of the text. In other words, the word's context is similar to the overall context of the text. This suggests that the word is closely related and contributes most to the main topic or theme of the text. On the other hand, a low cosine similarity score indicates that the word's embedding and the mean embedding are pointing in different directions. This suggests that the word's context does not align well with the overall context of the text and the word is less central to the main theme of the text. As the BERT tokenizer uses sub-word tokenization, so a single word can get tokenized into different tokens. To address this, we accumulate all the sub-words to form the original word and consider the average of all the sub-word token embeddings as the word embedding. As an example, we consider the word ``{\bng মোজাফফর}'' (``Mozaffar'' which is the name of a person). During tokenization, the word gets tokenized into three sub-word tokens: {\bng মোজা}, \#\#{\bng ফ} and \#\#{\bng ফর}. At first, we get the embedding for each of these three sub-tokens. We store the embedding of the first sub-token without ``\#\#'' prefix i.e. the token embedding of ``{\bng মোজা}''. Then, we check if a token has ``\#\#'' as the prefix. As long as a token has ``\#\#'' prefix, we continue to add the embeddings of the sub-word tokens. When all the sub-word tokens of a word are traversed, we stop the process and calculate the average. In summary, we consider the average of the sub-token embeddings (average of the embeddings of {\bng মোজা, \#\#ফ, \#\#ফর}) as the word embedding of ``{\bng মোজাফফর}''.

\subsubsection{Keywords Selection} 
After assigning importance scores to all words in a text, we use a threshold value which indicates the percentage of words to be extracted. For texts with $10$ or more words, we returned the top $60\%$ of words in an unordered manner. For texts with $5$ to $9$ words, we returned $70\%$ of the words, while for texts with four words or fewer, we returned $80\%$ of the words. There are two reasons behind such threshold values. First, texts with a smaller number of words inherently have less information available. Therefore, a higher threshold can be used to ensure that enough relevant words are captured to represent the content adequately. Second, shorter texts do not contain enough context to accurately determine the importance of individual tokens. By returning a higher percentage of words, the model compensates for the lack of contextual information and provide a broader representation of the content. However, it is possible that such threshold value on shorter texts can force the extractor to select some less important words. For example, we consider a short text in Bangla ``{\bng তাই সে কাজটি করেনি}'' (English Translation: ``That is why he did not do the work''). The Bangla text has only four words, while the corresponding translated English text has nine words. In this text, the two important words are ``{\bng কাজটি}'' (the work) and ``{\bng করেনি}'' (did not do). During text generation, the generation model is free to generate text using these two words with other diverse words. But including very few less important words i.e. ``{\bng তাই}'' (that is why) will not hurt the text generation performance instead will force the generation model to generate a text that contains less important words. By lowering the threshold value, we can control the selection of less important keywords.

\subsubsection{Baselines} 
For our proposed keyword extraction approach, we employed two publicly available pre-trained Bangla language models: \texttt{BanglaBERT} \citep{bhattacharjee-etal-2022-banglabert} and \texttt{Bangla-BERT-Base} \citep{Sagor_2020}. For performance comparison, we applied two other popular keyword extraction techniques: \texttt{TextRank} \citep{mihalcea-tarau-2004-textrank} and \texttt{YAKE} \citep{campos2020yake}. 
% The experimental settings and model descriptions can be found in Appendix \ref{app:A} and \ref{app:D} respectively.

\subsubsection{Evaluation Metrics} 
We utilized Mean Reciprocal Rank (MRR) \citep{voorhees1999trec}, Mean Average Precision (mAP) \citep{baeza1999modern} and Normalized Discounted Cumulative Gain (nDCG) \citep{jarvelin2017ir} scores to compare the performance of the keyword extractors. 
% Please see Appendix \ref{app:E} for details.

\subsubsection{Performance Analysis} 
We assessed the keyword extractors using the above mentioned evaluation metrics. To make comparisons, we conducted evaluations on the initial $10K$ data points from the test dataset, which contained manually extracted keywords. The findings are presented in Table \ref{table:extractors}. 
\begin{table}[h]
\centering
\resizebox{\columnwidth}{!}
{\begin{tabular}{ccccc} 
\hline
\textbf{Metrics} & \begin{tabular}[c]{@{}c@{}}\textbf{Bangla}\\\textbf{BERT}\end{tabular} & \begin{tabular}[c]{@{}c@{}}\textbf{Bangla}\\\textbf{BERT Base}\end{tabular} & \begin{tabular}[c]{@{}c@{}}\textbf{Text }\\\textbf{Rank}\end{tabular} & \textbf{YAKE}   \\ 
\hline
\textbf{MRR}     & \textbf{33.59}                                                         & 33.49                                                                       & 33.08                                                                 & 33.55           \\
\textbf{mAP}     & \textbf{33.56}                                                         & 32.43                                                                       & 31.60                                                                 & 31.63           \\
\textbf{nDCG}    & 33.81                                                                  & 31.11                                                                       & 27.01                                                                 & \textbf{37.91}  \\
\hline
\end{tabular}}
\caption{Comparative analysis among four keyword extraction methods. Results are expressed in percentages.}
\label{table:extractors}
\end{table}
It is evident from the table that both \texttt{BanglaBERT} and \texttt{Bangla-BERT-Base} exhibited similar performance across all metrics. However, \texttt{BanglaBERT} attained higher scores. They achieved slightly higher Mean Reciprocal Rank (MRR) and Mean Average Precision (mAP) compared to \texttt{TextRank} and \texttt{YAKE}. \texttt{TextRank} consistently performed slightly lower than the BERT-based models in terms of MRR and mAP, but notably lower in terms of nDCG. This suggests that \texttt{TextRank} does not capture the nuances of the dataset as effectively as BERT-based methods, especially in terms of ranking relevance. \texttt{YAKE} outperforms other techniques significantly in terms of nDCG, indicating that it is more effective in capturing the relevance of keywords in the dataset. However, it performs lower in terms of MRR and mAP compared to \texttt{BanglaBERT} and \texttt{Bangla-BERT-Base}. Based on the results, we employed \texttt{BanglaBERT} based keyword extractor (as our proposed) to automatically extract keywords from all the texts within the \textit{Bangla Key2Text} dataset.

%Selection of decoding technique table
\begin{table*}[h]
\centering
\resizebox{0.97\textwidth}{!}{
\begin{tabular}{c|cc|cc|cc|cc|cc} 
\hline
\multirow{2}{*}{\textbf{Metrics }} & \multicolumn{2}{c|}{\textbf{Top-p}} & \multicolumn{2}{c|}{\textbf{Beam }} & \multicolumn{2}{c|}{\textbf{Greedy }} & \multicolumn{2}{c|}{\textbf{Top-k}} & \multicolumn{2}{c}{\textbf{Top-p \& Top-k}}  \\ 
\cline{2-11}
                                   & \textbf{mT5}   & \textbf{bnT5}      & \textbf{mT5} & \textbf{bnT5}        & \textbf{mT5} & \textbf{bnT5}          & \textbf{mT5}   & \textbf{bnT5}      & \textbf{mT5}   & \textbf{bnT5}               \\ 
\hline
\textbf{BERT-S}                    & \textbf{91.07} & 91.51              & 90.73        & \uline{92.76}        & 89.02        & 92.45                  & 91.1           & 92.38              & 91.03          & 92.41                       \\
\textbf{ROUGE-1}                   & \textbf{60.63} & 56.34              & 59.6         & \uline{61.89}        & 50.84        & 61.36                  & 60.04          & 60.96              & 60.6           & 61.15                       \\
\textbf{ROUGE-L}                   & 45.18          & 43.23              & 44.02        & \uline{48.66}        & 37.61        & 47.62                  & \textbf{45.72} & 47.19              & 45.02          & 47.42                       \\
\textbf{BLEU-3}                    & 11.59          & 11.36              & 10.78        & \uline{16.45}        & 6.55         & 15.37                  & 11.44          & 14.91              & \textbf{12.59} & 15.04                       \\
\textbf{BLEU-4}                    & 5.04           & 5.06               & 4.44         & \uline{8.16}         & 2.45         & 7.61                   & 5.04           & 7.36               & \textbf{5.07}  & 7.46                        \\
% \textbf{METEOR}                    & \textbf{36.46} & 33.26              & 35.32        & \uline{39.17}        & 27.75        & 38.19                  & 36.32          & 37.73              & 36.42          & 37.89                       \\
\textbf{WER}                       & \textbf{75.3}  & 79.9               & 76.96        & \uline{71.82}        & 86.46        & 73.6                   & 76.31          & 74.2               & 76.18          & 74.06                       \\
\textbf{WIL}                       & \textbf{83.47} & 84.81              & 84.57        & \uline{80.03}        & 89.38        & 80.99                  & 84.09          & 81.34              & 84.61          & 81.27                       \\
\hline
\end{tabular}}
\caption{Text generation performance comparison between the \textit{mT5} and \textit{BanglaT5 (bnT5)} models on initial $10K$ test data using various decoding techniques. Results are presented as percentages. BERT-S represents BERTScore. \textbf{Bold} values indicate the maximum scores achieved by the \textit{mT5} model, while \underline{underlined} values indicate the maximum scores achieved by the \textit{BanglaT5} model.}
\label{tab:decoding-performance}
\end{table*}

\section{Text Generation}\label{generation}
Transformer \citep{vaswani2017attention} based pre-trained sequence-to-sequence models, when fine-tuned on natural language generation tasks, can yield exceptional results \citep{rothe2020leveraging}. We fine-tuned two pre-trained sequence-to-sequence T5 \citep{raffel2020exploring} based models for Bangla text generation task from keywords using our \textit{Bangla Key2Text} dataset. 
% The T5 model architecture consists of a stack of transformer encoder-decoder layers. Each layer in the T5 model contains multiple self-attention heads and feed-forward neural networks. The T5 model takes a sequence of tokens as input, which are obtained by tokenizing the keywords. The input tokens are then embedded into continuous vector representations, which capture the semantic meaning of the tokens. The encoder utilizes self-attention and feed-forward neural networks to capture dependencies and relationships between words and phrases. The decoder incorporates an additional cross-attention mechanism, attending to both the encoded representation and previously generated words during the decoding process. 
% Fine-tuning a pre-trained language model with $2$ million data takes significant amount of time and computational resources which explains why Bangla is still considered a low resource language. Therefore, we provide the two fine-tuned T5 models as benchmarks on Bangla text generation task from provided keywords. 
% We have made the dataset and the two fine-tuned models publicly available in order to facilitate research on keywords to text generation task.

\subsection{Baselines} 
To establish baseline performance on the Bangla Key2Text dataset, we fine-tune two transformer-based encoder–decoder models, \texttt{BanglaT5} \citep{bhattacharjee-etal-2023-banglanlg} and \texttt{(mT5)} \citep{hasan-etal-2021-xl}, as task-specific baselines for keyword-to-text generation. We additionally evaluate several zero-shot open-source large language models (LLMs) to examine how general-purpose models perform without task-specific fine-tuning. We also evaluated zero-shot capabilities of several 4-bit quantized (due to computational resource limitations) open-source large language models. These models include Phi-3.5 mini (3.8B), mGPT (1.3B), gemma-2 (2B), Mistral (7B), Ministral (8B), Llama 3.1 (8B), Qwen-2.5 (7B), Bangla-llama-2 (7B).
% These include \texttt{microsoft/Phi-3.5-mini-instruct}, \texttt{ai-forever/mGPT}, \texttt{google/gemma-2-2b}, \texttt{mistralai/Mistral-7B-Instruct-v0.3}, \texttt{mistralai/Ministral-8B-Instruct-2410}, \texttt{meta-llama/Meta-Llama-3.1-8B-Instruct}, \texttt{Qwen/Qwen2.5-7B-Instruct}, and \texttt{BanglaLLM/bangla-llama-7b-instruct-v0.1}, a Bangla-aligned variant of LLaMA-2. 
These models were evaluated using their default inference settings without any fine-tuning. 

\subsection{Experimental Settings} 
We fine-tuned both the T5 models for two epochs. \texttt{mT5} was fine-tuned using an NVIDIA GeForce RTX 3090 24GB GPU and \texttt{BanglaT5} was fine-tuned using an NVIDIA GeForce RTX 3060 12GB GPU. All models were implemented using \texttt{PyTorch} \citep{paszke2019pytorch} framework. During the fine-tuning process, the data was organized into batches, with batch size $2$ for \texttt{mT5} and batch size $1$ for \texttt{BanglaT5}. We utilized a learning rate of $1e-4$, employed the \textit{AdamW} optimizer with a linear warm up of $500$ steps and `OneCycleLR' learning rate scheduler. The scheduler was configured with various parameters including a division factor of $10$, a final division factor of $100$, a starting percentage of $10\%$ and a linear annealing strategy. To accommodate the model's token limitations, we truncated the input sequences to $512$ tokens and added PAD tokens where necessary. Considering the average number of keywords and the average number of words in a text in our \textit{Bangla Key2Text} dataset, we set the max length for text generation to $64$ tokens. During inference, we experimented with different decoding techniques i.e. greedy decoding, beam search with beam size $2$, top-k sampling with $k=50$ and top-p sampling with $p=0.95$. The repetition penalty and length penalty were set to $2.5$ and $\alpha=1.0$ respectively. For 4-bit quantized LLMs, we utilized model-specific default text generation parameters during inference.

\subsection{Evaluation Metrics}
We assessed the performance of the text generation models using several standard natural language generation (NLG) metrics that includes BERTScore \citep{zhang2019bertscore}, ROUGE \citep{lin2004rouge}, BLEU \citep{papineni2002bleu}, WER \citep{ali2018word}, and WIL\footnote{\url{https://pytorch.org/torcheval/stable/generated/torcheval.metrics.WordInformationLost.html}}. 
% More details can be found in Appendix \ref{app:E}. 
We also conducted human evaluations on the quality of the generated texts by native Bangla speakers.

\subsection{Results and Discussion}\label{text-gen-results}
In this section, we address the performance of the text generation models based on extensive experimentation.

\noindent \textbf{(a) Selection of decoding technique.}
We tested five decoding methods - \textit{greedy}, \textit{beam search}, \textit{top-k sampling}, \textit{top-p sampling}, and a combination of \textit{top-p \& top-k} on the $10K$ evaluation set for both \texttt{mT5} and \texttt{BanglaT5} to identify the most effective approach, then applied the best method to the full $100K$ test set. Results (Table \ref{tab:decoding-performance}) show that \texttt{mT5} performed best with \textit{top-p sampling}, while \texttt{BanglaT5} achieved its highest scores with \textit{beam search}.

\noindent\textbf{(b) Performance of the fine-tuned models.}
We evaluated the proposed fine-tuned models on the $100K$ test set using the specified metrics (Table \ref{table:4.2}), applying \textit{top-p sampling} for \texttt{mT5} and \textit{beam search} for \texttt{BanglaT5}.
\begin{table}[h]
\centering
\resizebox{0.75\columnwidth}{!}{
    \begin{tabular}{ccc} 
    \hline
    \textbf{Metrics} & \textbf{mT5}   & \textbf{BanglaT5}  \\ 
    \hline
    \textbf{BERTScore}  & 91.06          & \textbf{91.51}     \\
    \textbf{ROUGE-1} & \textbf{60.62} & 56.36              \\
    \textbf{ROUGE-L} & \textbf{45.15} & 43.20              \\
    \textbf{BLEU-3}  & \textbf{11.68} & 11.37              \\
    \textbf{BLEU-4}  & 5.06           & \textbf{5.07}      \\
    % \textbf{METEOR}  & \textbf{36.43} & 33.18              \\
    \textbf{WER}     & \textbf{75.34} & 79.98              \\
    \textbf{WIL}     & \textbf{83.47} & 84.85              \\
    \hline
    \end{tabular}
}
\caption{Text generation performance comparison between the \textit{mT5} and \textit{BanglaT5} models on the $100K$ test set. Results are presented as percentages.}
\label{table:4.2}
\end{table}
BERTScore (F1) reached $91.06\%$ for \texttt{mT5} and $91.51\%$ for \texttt{BanglaT5}, reflecting strong semantic alignment between generated and reference texts. ROUGE-L was lower - $45.15\%$ and $43.20\%$, respectively - indicating room for improvement in capturing longer sequences. BLEU-3 and BLEU-4 were also relatively low, showing limited higher-order n-gram similarity. Word Error Rate (WER) was high ($75.34\%$ for \texttt{mT5}, $79.98\%$ for \texttt{BanglaT5}), while Word Information Lost (WIL) scores ($83.47\%$ and $84.85\%$) suggest notable information loss at the word level.\\

We observe relatively high BERTScore values alongside high WER. This occurs because multiple valid sentences can be generated from the same keywords, leading to lexical differences from the reference text; BERTScore captures semantic similarity while WER penalizes surface-level mismatches.\\

\noindent\textbf{(c) Impact of Temperature.} We studied how the temperature parameter ($\tau$) affects text generation in \texttt{mT5} and \texttt{BanglaT5} using the initial $10K$ evaluation set and three values ($\tau=0.3,0.7,1.0$). Performance at $\tau=1.0$ is shown in Table \ref{tab:decoding-performance}, while results for $\tau=0.3$ and $\tau=0.7$ are in Table \ref{tab:temperature-test}. For \texttt{mT5}, we used top-p sampling; for \texttt{BanglaT5}, greedy and beam search - since these worked best at $\tau=1.0$. The results show that for \texttt{BanglaT5}, beam search with $\tau=0.7$ outperforms $\tau=0.3$ across all metrics, with a similar trend for greedy decoding. Likewise, in \texttt{mT5}, lowering $\tau$ from $0.7$ to $0.3$ reduces performance.

%temperature test table
\begin{table}[h]
\centering
\resizebox{1\columnwidth}{!}{
\begin{tabular}{c|cc|cc|cc} 
\hline
\multirow{2}{*}{\textbf{Metrics}} & \multicolumn{2}{c|}{\textbf{BanglaT5 (Beam)}} & \multicolumn{2}{c|}{\textbf{BanglaT5 (Greedy)}} & \multicolumn{2}{c}{\textbf{mT5 (Top-p)}}  \\ 
\cline{2-7}
                                  & \textbf{$\tau=0.3$} & \textbf{$\tau=0.7$}           & \textbf{$\tau=0.3$} & \textbf{$\tau=0.7$}             & \textbf{$\tau=0.3$} & \textbf{$\tau=0.7$}       \\ 
\hline
\textbf{BERTScore}                   & 88.6             & 92.7                       & 91.95            & 92.38                        & 83.8             & 90.5                   \\
\textbf{ROUGE-1}                  & 47.29            & 62.26                      & 58.68            & 60.96                        & 29.35            & 59.22                  \\
\textbf{ROUGE-L}                  & 39.07            & 49.13                      & 45.12            & 47.22                        & 24.35            & 44.29                  \\
\textbf{BLEU-3}                   & 11.93            & 16.66                      & 13.14            & 14.97                        & 5.62             & 11.09                  \\
\textbf{BLEU-4}                   & 5.93             & 8.4                        & 6.23             & 7.35                         & 2.15             & 4.69                   \\
% \textbf{METEOR}                   & 31.75            & 39.7                       & 35.41            & 37.7                         & 22.61            & 35.9                   \\
\textbf{WER}                      & 136.99           & 70.92                      & 77.09            & 74.2                         & 110.2            & 75.45                  \\
\textbf{WIL}                      & 83.56            & 79.33                      & 83.22            & 81.37                        & 91.21            & 83.63                  \\
\hline
\end{tabular}}
\caption{Effect of different temperature settings on the text generation quality of the fine-tuned \textit{mT5} and \textit{BanglaT5} models.}
\label{tab:temperature-test}
\end{table}

\noindent\textbf{(d) Generalization performance on unseen and dialect data}.
To evaluate generalization, we collected $1000$ Bangla news articles from Prothom Alo\footnote{\url{https://www.prothomalo.com/}} that were not included in the training corpus. Keywords were extracted both manually and automatically using our proposed extractor, and generated texts were evaluated using the same metrics as the main test set (results in Table \ref{tab:external-dataset}).
%Result on external news dataset
\begin{table}[h]
\centering
\resizebox{\columnwidth}{!}
{\begin{tabular}{c|cc|cc} 
\hline
\multirow{2}{*}{\textbf{Metrics}} & \multicolumn{2}{c|}{~\textbf{Proposed Extractor}} & \multicolumn{2}{c}{\textbf{~Human Extracted}}  \\ 
\cline{2-5}
                                  & \textbf{mT5}   & \textbf{BanglaT5}                & \textbf{mT5} & \textbf{BanglaT5}               \\ 
\hline
\textbf{BERT-S}                   & 91.25          & \textbf{91.85}                   & 90.66        & \textbf{92.14}                  \\
\textbf{ROUGE-1}                  & \textbf{63.51} & 59.30                            & 60.89        & \textbf{61.93}                  \\
\textbf{ROUGE-L}                  & \textbf{50.03} & 48.26                            & 48.11        & \textbf{50.75}                  \\
\textbf{BLEU-3}                   & \textbf{12.12} & 10.86                            & 9.55         & \textbf{12.29}                  \\
\textbf{BLEU-4}                   & \textbf{5.44}  & 4.33                             & 3.78         & \textbf{5.02}                   \\
% \textbf{METEOR}                   & \textbf{38.35} & 34.48                            & 36.06        & \textbf{37.21}                  \\
\textbf{WER}                      & \textbf{72.33} & 75.70                            & 73.79        & \textbf{71}                     \\
\textbf{WIL}                      & \textbf{81.68} & 83.95                            & 83.37        & \textbf{81.53}                  \\
\hline
\end{tabular}}
\caption{Text generation performance comparison between the \textit{mT5} and \textit{BanglaT5} models on the unseen $1K$ news data. Results are presented as percentages. BERT-S represents BERTScore.}
\label{tab:external-dataset}
\end{table}
Our findings show \texttt{mT5} performs best with proposed extractor, while \texttt{BanglaT5} performs best with human-extracted keywords. The exact match between manual and automatic keywords is $74.10\%$. To assess text generalization, we collected Bangla regional dialect keywords and generated texts with our fine-tuned models.
% (examples in Appendix \ref{app:C} Table \ref{tab:dialect-data})
The models could produce dialectal texts from the keywords, but quality was limited which is expected since neither \texttt{mT5} nor \texttt{BanglaT5} was fine-tuned on dialect data. Dialects pose greater challenges than standard Bangla due to varied grammar, vocabulary, idioms, and contextual sensitivities. These factors necessitate dialect-specific data and fine-tuning, which we identify as a promising direction for future work. This dialect evaluation is exploratory and qualitative, as no standardized Bangla dialect dataset exists for keyword-to-text generation.\\

\noindent\textbf{(e) Human Evaluation}. 
We conducted a human evaluation with three expert annotators.
 % (a graduate student and two academic specialists; recruitment details in Appendix \ref{app:H})
They reviewed $1000$ randomly selected keyword–text pairs from the test set, along with outputs from both models, using our hosted inference API. Each annotator independently assessed all $1000$ texts for coherence and relevance to the input keywords, considering semantic alignment with the original text and keyword presence. 
\begin{table}[h]
\centering
\resizebox{0.9\columnwidth}{!}
{\begin{tabular}{cccc} 
\hline
\textbf{Model}            & \textbf{~ Category~~} & \begin{tabular}[c]{@{}c@{}}\textbf{Kappa}\\\textbf{Score}\end{tabular} & \textbf{Average}       \\ 
\hline
\multirow{2}{*}{\textbf{mT5}}      & Relevant              & 0.91                                                                   & \multirow{2}{*}{0.87}  \\ 
\cline{2-3}
                          & Irrelevant            & 0.83                                                                   &                        \\ 
\hline
\multirow{2}{*}{\textbf{BanglaT5}} & Relevant              & 0.87                                                                   & \multirow{2}{*}{0.84}  \\ 
\cline{2-3}
                          & Irrelevant            & 0.81                                                                   &                        \\
\hline
\end{tabular}}
\caption{Fleiss' kappa score for inter-annotator agreement on \textit{relevant/partially relevant} and \textit{irrelevant} categories.}
\label{tab:kappa-score}
\end{table}
They labeled each pair as \textit{relevant}, \textit{partially relevant}, or \textit{irrelevant}, framing the task as a data-labeling problem with distinct classes. Inter‑annotator agreement was measured using Fleiss’ kappa \citep{fleiss1971measuring}, yielding scores of $0.87$ for \texttt{mT5} and $0.84$ for \texttt{BanglaT5} (Table \ref{tab:kappa-score}), indicating substantial agreement. Although only three annotators were involved, this setup follows common practice in NLG evaluation and the high Fleiss’ κ scores indicate reliable agreement.

\begin{table*}[h]
\centering
\resizebox{\textwidth}{!}{
\begin{tabular}{ccccccccccc} 
\hline
\textbf{Metrics} & \textbf{BanglaT5}  & \textbf{mT5} & \textbf{Phi-3.5} & \textbf{mGPT} & \textbf{Gemma-2B} & \textbf{Mistral} & \textbf{Ministral} & \textbf{LLaMA-3.1} & \textbf{LLaMA-2} & \textbf{Qwen}  \\ 
\hline
\textbf{WER}      & \textbf{71.82} & 76.96    & 96.19      & 103.58      & 105.09      & 104.55      & 107.11 & 99.76 & 125.3 & 101.07     \\
\textbf{WIL}      & \textbf{80.03}  & 84.57   & 97.86  & 99.56      & 99.69      & 99.83      & 99.58 & 99.26 & 99.6  & 99.66      \\
\textbf{BERTScore} & \textbf{92.76} & 90.73   & 77.41      & 79.43      & 78.57      & 78.16      & 79.03 & 77.78 & 84.11 & 78.61     \\
\textbf{ROUGE-1}  & \textbf{61.89} & 59.6  & 11.95      & 4.99      & 4.08      & 1.82      & 5.58 & 5.43 & 7.34 & 3.35      \\
\textbf{ROUGE-L}  & \textbf{48.66}  & 44.02   & 10.31      & 4.37      & 3.62      & 1.7      & 4.6 & 4.76 & 6.52 & 2.93      \\
\textbf{BLEU-3}  & \textbf{16.45} & 10.78  & 2.32e-03      & 3.58e-05      & 1.23e-04      & 2.99e-05      & 3.02e-04 & 9.70e-06 & 8.53e-05 & 1.11e-105      \\
% \textbf{BLEU-4}  & \textbf{8.16}  & 4.46   & 2.84e-80      & 5.68e-82      & 4.11e-05      & 4.08e-82      & 6.69e-05 & 1.18e-82 & 6.52 & 6.51e-158      \\
\hline
\end{tabular}
}
\caption{Performance comparison of open-source large language models (LLMs) with BanglaT5 across various metrics. \textbf{Bold} values indicate the best scores. LLaMA-2 here represents the Bangla-llama-2 model.}
\label{tab:llm-model-comparison}
\end{table*}

\begin{table*}[t]
\centering
\resizebox{\textwidth}{!}{
\begin{tabularx}{\textwidth}{XX}

% \textbf{Keywords to Text generation}                                                           & \textbf{English Translation}                                                                                                 \\
\hline
\textbf{Keys: [B]} {\bng শীতের, কম, আগে, সম্ভাবনা}                                                                & \textbf{{[}E]} winter, slim, before, chance                                                                                \\ 
\textbf{Original Text: [B]} {\bng  শীতের আগে সেই সম্ভাবনা কম।} & \textbf{{[}E]} That chance is slim before winter.          \\ 
\textbf{mT5: [B]} {\bng শীতের আগে এমন সম্ভাবনা কম।}                                                   & \textbf{{[}E]} Such chance is slim before winter.                                                               \\ 
\textbf{BanglaT5: [B]} {\bng শীতের আগে আসার সম্ভাবনা কম।}                                                   & \textbf{{[}E]} The chance of arriving before winter is slim.                                                               \\ \hdashline
\textbf{Reordered Keys: [B]} {\bng সম্ভাবনা, কম, আগে, শীতের}                                                   & \textbf{{[}E]} chance, slim, before, winter                                                       \\ 
\textbf{mT5: [B]} {\bng শীতের আগে এমন সম্ভাবনা কম।}                                                   & \textbf{{[}E]} Such chance is slim before winter.                                                               \\ 
\textbf{BanglaT5: [B]} {\bng শীতের আগে এ সম্ভাবনা কম।}                                                   & \textbf{{[}E]} The chance is slim before winter.                                                               \\ \hline
\textbf{Keys: [B]} {\bng পেল, খবর, কোত্থেকে, বানোয়াট}                                                                & \textbf{{[}E]} get, news, where, fake                                                                                \\ 
\textbf{Original Text: [B]} {\bng তারা কোত্থেকে পেল এসব বানোয়াট খবর? } & \textbf{{[}E]} Where did they get these fake news?          \\ 
\textbf{mT5: [B]} {\bng কোত্থেকে বানোয়াট খবর পেল তাঁরা?}                                                   & \textbf{{[}E]} Where did they get fake news?                                                               \\ 
\textbf{BanglaT5: [B]} {\bng কোত্থেকে বানোয়াট খবর পেল তাঁরা?}                                                   & \textbf{{[}E]} Where did they get fake news?                                                               \\ \hdashline
\textbf{Reordered Keys: [B]} {\bng বানোয়াট, খবর, পেল, কোত্থেকে}                                                   & \textbf{{[}E]} fake, news, get, where                                                               \\ 
\textbf{mT5: [B]} {\bng কোত্থেকে বানোয়াট খবর পেল তাঁরা?}                                                   & \textbf{{[}E]} Where did they get fake news?                                                               \\ 
\textbf{BanglaT5: [B]} {\bng বানোয়াট খবর কোত্থেকে পেল সে?}                                                   & \textbf{{[}E]} Where did he get fake news?                                                              \\ \hline
\textbf{Keys: [B]} {\bng ছেলেটা, ঢাকায়, আসে, গ্রামের}                                                                & \textbf{{[}E]} boy, Dhaka, comes, village                                                                                \\ 
\textbf{Original Text: [B]} {\bng গ্রামের সহজ-সরল ছেলেটা ঢাকায় আসে। } & \textbf{{[}E]} The simple village boy comes to Dhaka.           \\ 
\textbf{mT5: [B]} {\bng ছেলেটা গ্রামের বাড়ি থেকে ঢাকায় আসে।}                                                   & \textbf{{[}E]} The boy comes to Dhaka from his village home.                                                               \\ 
\textbf{BanglaT5: [B]} {\bng গ্রামের ছেলেটা ঢাকায় আসে না।}                                                   & \textbf{{[}E]} The village boy does not come to Dhaka.                                                             \\ \hdashline          
\textbf{Reordered Keys: [B]} {\bng গ্রামের, ছেলেটা, আসে, ঢাকায়}                                                   & \textbf{{[}E]} village, boy, comes, Dhaka                                                             \\            
\textbf{mT5: [B]} {\bng ছেলেটা গ্রামের বাড়ি থেকে ঢাকায় আসে।}                                                   & \textbf{{[}E]} The boy comes to Dhaka from his village home.                                                              \\             
\textbf{BanglaT5: [B]} {\bng গ্রামের ছেলেটা ঢাকায় আসে না।}                                                   & \textbf{{[}E]} The village boy does not come to Dhaka.                                                                 \\ \hline            
\end{tabularx}
}
\caption{Few examples of generated texts by the fine-tuned \textit{mT5} and \textit{BanglaT5} models, showcasing the impact of different input keyword orders. Each row represents a distinct example, with two variations of keyword order provided. These examples are drawn from the test dataset. \textbf{B} represents the keywords or texts in Bangla and \textbf{E} represents their corresponding English translation.}
\label{tab:keyword-order}
\end{table*}

\noindent\textbf{(f) Performance comparison with LLMs.} 
To evaluate how well general-purpose LLMs handle Bangla keyword-to-text generation, we tested eight open-source, 4-bit quantized models (due to resource constraints) in a zero-shot setting on the $10K$ evaluation set using their default decoding settings. As shown in Table \ref{tab:llm-model-comparison}, \texttt{BanglaT5} achieved the best overall results across all metrics, with \texttt{mT5} following closely - particularly strong on semantic alignment (BERTScore). In contrast, the other models, including \texttt{Phi‑3.5}, \texttt{mGPT}, \texttt{Gemma}, \texttt{Qwen}, and \texttt{Mistral}, performed poorly, with low ROUGE and high WER/WIL scores, reflecting outputs that were often short, disfluent, or misaligned with the keywords. Even \texttt{Bangla-LLaMA-2} and \texttt{LLaMA-3.1}, despite their larger Bangla vocabularies and instruction tuning, offered only marginal improvements - echoing earlier findings \citep{kabir2023benllmeval, mahfuz-etal-2025-late}. Note that these LLMs are evaluated in a zero-shot setting without task-specific fine-tuning; our goal is to illustrate the importance of task-specific training for low-resource languages such as Bangla. These results suggest that for low-resource languages like Bangla, task-specific fine-tuning remains crucial, as general-purpose LLMs - even when compressed - struggle to generate coherent and keyword-faithful text.

\noindent\textbf{(g) Effect of number of keywords during generation}.
In the \textit{Bangla Key2Text} dataset, texts average $9.381$ keywords and $15.142$ words. Based on these values, we set the maximum generation length to $64$ tokens, though this is adjustable. Trial and error by adding and removing keywords showed that generating longer sequences (65-100 tokens) requires more keywords - typically $10$-$15$ produce satisfactory multi-sentence outputs. This is because more keywords prompt the models to add additional context words, leading to richer texts.
% (see Appendix \ref{app:C} Table \ref{tab:single-line-test})
We also tested both models with only a single keyword and found they produced very short but complete texts.
% (see Appendix \ref{app:C} Table \ref{tab:single-key})

\noindent\textbf{(h) Constrained Beam Search.}
Constrained beam search \citep{hokamp-liu-2017-lexically} enforces the inclusion of specified tokens at each generation step. In our experiments, it successfully ensured missing keywords appeared in the output but had limitations. We used a two-stage brute-force approach: first generating text with standard beam search, then checking which keywords were missing, and finally regenerating the text with constrained beam search to enforce their inclusion while keeping other parameters unchanged. However, forcing all keywords in this way reduced text quality, as compelling the model to include one keyword often led to the exclusion of others.

\noindent\textbf{(i) Cross-Lingual Transfer.}
During generation, we observed cross-lingual transfer \citep{schuster-etal-2019-cross-lingual}, where both \texttt{mT5} and \texttt{BanglaT5} produced Bangla text even when given mixed Bangla‑English keywords. 
% (see Appendix \ref{app:C} Table \ref{tab:cross-lingual-transfer})
Fine‑tuned T5 models specialize in their training language, while multilingual models like \texttt{mT5} leverage cross‑lingual understanding. Since both models were fine‑tuned on Bangla, they favor Bangla outputs. For \texttt{mT5}, this means interpreting English keywords, mapping them to Bangla, and generating consistent Bangla text. \texttt{BanglaT5}, however, sometimes retains English words in the output, though their placement remains semantically appropriate. Future research could further explore this cross-lingual behavior and develop better strategies to control mixed generation.

\noindent\textbf{(j) Sensitivity to Keyword Order}
We conducted manual testing on the text generation behavior of the fine-tuned models. We randomly selected few keyword-text pairs from the test dataset. For each pair, we shuffled the keyword order (both randomly and preserving the original appearance order in the input text) and observed the generated texts from both models. We present some examples in Table \ref{tab:keyword-order}. Our focus was on assessing the impact of using different variations of input keyword order on the text generation process. The examples illustrate that altering the input keyword order results in minor differences in the generated output, highlighting the model's sensitivity to keyword order. However, in all the examples, although the syntactic structures of the generated texts differ, their semantic meaning remains consistent compared to the original text. In most cases, even with changes in keyword order, the generated text maintains both syntactic and semantic coherence. This observation aligns with the high BERTScores $91.06\%$ (\texttt{mT5}) and $91.51\%$ (\texttt{BanglaT5}) indicating substantial semantic correctness between the reference and generated texts. Nevertheless, it is essential to quantify the sensitivity of the text generation models to changes in keyword order using proper evaluation metrics and delve deeper into analyzing the effects. We leave this task open for future work.

\section{Conclusion and Future Works}
In this work, we introduced \textit{Bangla Key2Text}, a large-scale dataset of $2.6$ million Bangla keyword-text pairs designed for keyword-to-text generation. We developed a BERT-based keyword extractor and fine-tuned \texttt{mT5} and \texttt{BanglaT5}, demonstrating through extensive evaluation that they generate coherent, keyword-faithful Bangla text. Our experiments - covering zero-shot comparisons with multiple compressed LLMs, human evaluations, unseen data, and dialectal inputs - show that task-specific fine-tuning of smaller language models significantly outperforms general-purpose models in this low-resource setting. Moving forward, we plan to fine-tune LLMs and monolingual models \citep{bhattacharjee-etal-2023-banglanlg}, explore few-shot learning, and further investigate cross-lingual keyword-to-text generation \citep{chi2020cross}.

\section{Limitations}
We observed that the generated texts occasionally failed to incorporate all provided keywords, leading to incomplete representation of the intended content. This highlights the need for more effective decoding strategies, such as improved constrained generation. Also, most extracted keywords used for fine-tuning were not lemmatized, which sometimes caused fluency and grammatical issues when generating text from lemmatized or morphologically varied keywords. Additionally, since the dataset is derived primarily from news articles, models trained on it may generalize less effectively to conversational or informal Bangla. Another limitation is that our models lack mechanisms to detect or filter harmful content such as slang, hate speech, or adult text, raising ethical and safety concerns. Moreover, while our fine-tuned models significantly outperform general-purpose LLMs, their performance on dialectal and cross-lingual inputs remains limited, and the cross-lingual transfer behavior observed warrants deeper investigation. Finally, as our work focused on 4-bit quantized and fine-tuned models, the impact of higher-precision models or more extensive fine-tuning remains unexplored. Addressing these limitations could improve the reliability, safety, and generalizability of Bangla keyword-to-text generation.

\section{Ethical Considerations}
Our dataset, Bangla Key2Text, was compiled exclusively from publicly available Bangla news articles and online text sources. All materials were collected and used under fair-use provisions for research and educational purposes. No personally identifiable information (PII) or user-generated private data were included. The dataset and models are intended strictly for academic and non-commercial research under open licensing guidelines.

\nocite{Eco:1990,Martin-90,Chercheur-94,CastorPollux-92,bs-2570-manual,bs-2570-techreport,chomsky-73,hoel-71-portion,hoel-71-whole,singer-whole,Jespersen:1922,Superman-Batman-Catwoman-Spiderman-00}
% \section{Bibliographical References}\label{sec:reference}

\bibliographystyle{lrec2026-natbib}
\bibliography{lrec2026-example}

% \clearpage

\section*{Appendix}

\appendix

\section{Experimental Settings}\label{app:A}

\subsection{Keyword Extraction} 
The BERT based models were utilized for inference using their default hyper-parameters in the keyword extraction phase. No fine-tuning was performed. The tokenized text is passed through the pre-trained model to obtain the \textit{last hidden state} as the token embeddings, which had a shape of $512\times768$. By computing the average of the embeddings across the $512$ tokens, we derived a mean embedding of shape $768$.

\subsection{Text Generation} 
We fine-tuned both the T5 models for two epochs. \verb|mT5| was fine-tuned using an NVIDIA GeForce RTX 3090 24GB GPU and \verb|BanglaT5| was fine-tuned using an NVIDIA GeForce RTX 3060 12GB GPU. All models were implemented using \verb|PyTorch| \citep{paszke2019pytorch} framework. During the fine-tuning process, the data was organized into batches, with batch size $2$ for \verb|mT5| and batch size $1$ for \verb|BanglaT5|. We utilized a learning rate of $1e-4$, employed the \textit{AdamW} optimizer with a linear warm up of $500$ steps and `OneCycleLR' learning rate scheduler. The scheduler was configured with various parameters including a division factor of $10$, a final division factor of $100$, a starting percentage of $10\%$ and a linear annealing strategy. To accommodate the model's token limitations, we truncated the input sequences to $512$ tokens and added PAD tokens where necessary. Considering the average number of keywords and the average number of words in a text in our \textit{Bangla Key2Text} dataset, we set the max length for text generation to $64$ tokens. During inference, we experimented with different decoding techniques i.e. greedy decoding, beam search with beam size $2$, top-k sampling with $k=50$ and top-p sampling with $p=0.95$. The repetition penalty and length penalty were set to $2.5$ and $\alpha=1.0$ respectively. For 4-bit quantized LLMs, we utilized model-specific default text generation parameters during inference.

\section{Model Descriptions}\label{app:B}

\subsection{Keywords Extraction}
\textbf{(a) BanglaBERT.} A BERT model \citep{bhattacharjee-etal-2022-banglabert} fine-tuned for performing Bangla downstream tasks. The model checkpoint\footnote{\url{https://huggingface.co/csebuetnlp/banglabert}} can be accessed through the Hugging Face Transformers Library \citep{wolf2020transformers}. This discriminator model is pre-trained using ELECTRA \citep{clark2020electra} approach. It is comprised of a total of $110$ million parameters, with $12$ hidden layers and $12$ attention heads. The embedding size of the model is $768$ and the tokenizer associated with it has a vocabulary size of $32,000$.\\

\noindent \textbf{(b) Bangla-BERT-Base.} Another BERT model \citep{Sagor_2020} fine-tuned for performing Bangla downstream tasks. We fine-tuned the model checkpoint\footnote{\url{https://huggingface.co/sagorsarker/bangla-bert-base}} available on Hugging Face. It has the same embedding size as well as the same number of parameters, hidden layers, attention heads as the \verb|BanglaBERT| model. However, the tokenizer associated with it has a vocabulary size of $102,025$.\\

\noindent \textbf{(c) TextRank.} An algorithm \citep{mihalcea-tarau-2004-textrank} for text processing that is based on the \verb|PageRank| \citep{b16} algorithm used by Google for ranking web pages and websites. For keyword extraction, \verb|TextRank| identifies words in the text that are central and appear more often in the context of other words. The main idea is that words or phrases in a text are represented as nodes in a graph, and the connections between them are based on their co-occurrence in a certain window of words or based on other similarity measures. It works by assigning a score to each node in the graph based on the principle that a node is important if it is linked to by other important nodes. The algorithm iteratively calculates these scores until the process converges, meaning that the scores do not change significantly with further iterations. We employed a pre-trained \verb|BanglaBERT| model checkpoint\footnote{\url{https://huggingface.co/csebuetnlp/banglabert_generator}} from Hugging Face to compute the edge weights through cosine distance measurements among the text embeddings.\\

\noindent \textbf{(d) YAKE.} \textbf{Y}et \textbf{A}nother \textbf{K}eyword \textbf{E}xtractor \citep{campos2020yake} operates as an unsupervised automatic keyword extraction technique, relying on statistical features extracted from individual documents to identify the most significant keywords. The system does not require training on specific documents, nor does it rely on dictionaries or external corpora. It is publicly available as a Python library\footnote{\url{https://github.com/LIAAD/yake}}.

\subsection{Text Generation}
\noindent \textbf{(a) mT5.} A multilingual T5 model \citep{hasan-etal-2021-xl} fine-tuned extensively with abstractive summarization data covering 44 different languages. We trained the model checkpoint\footnote{\url{https://huggingface.co/csebuetnlp/mT5_multilingual_XLSum}} available on the Hugging Face Transformers Library \citep{wolf2020transformers} which is a fine-tuned version of the base \verb|mT5| model \citep{xue-etal-2021-mt5}. It contains a total of $600$ million parameters and consists of $12$ layers in both the encoder and decoder components. Each layer is equipped with $12$ attention heads. The tokenizer associated with this model has a vocabulary size of $250,112$.\\

\noindent \textbf{(b) BanglaT5.} A sequence-to-sequence \verb|T5| \citep{bhattacharjee-etal-2023-banglanlg} model that has been pre-trained using the span corruption objective with extensive Bangla texts to perform Bangla NLG tasks. The pre-trained model checkpoint\footnote{\url{https://huggingface.co/csebuetnlp/banglat5}} is available on the Hugging Face Library. It has the same number of encoder and decoder layers as the \verb|mT5|. However, it consists of $247$ million parameters and has a vocabulary of size $32,128$.\\

\noindent \textbf{(c) Large Language Models.} We additionally evaluated eight open-source, 4-bit quantized large language models in a zero-shot setting on the $10K$ evaluation set. These included \texttt{Phi-3.5-mini-instruct}\footnote{\url{https://huggingface.co/microsoft/Phi-3.5-mini-instruct}} (3.8B parameters), \texttt{mGPT}\footnote{\url{https://huggingface.co/ai-forever/mGPT}} (1.3B parameters), \texttt{Gemma-2B}\footnote{\url{https://huggingface.co/google/gemma-2-2b}} (2B parameters), \texttt{Mistral-7B}\footnote{\url{https://huggingface.co/mistralai/Mistral-7B-Instruct-v0.3}} (7B parameters), \texttt{Ministral-8B}\footnote{\url{https://huggingface.co/mistralai/Ministral-8B-Instruct-2410}} (8B parameters), \texttt{LLaMA-3.1--8B}\footnote{\url{https://huggingface.co/meta-llama/Meta-Llama-3.1-8B-Instruct}} (8B parameters), \texttt{Qwen2.5-7B}\footnote{\url{https://huggingface.co/Qwen/Qwen2.5-7B-Instruct}} (7B parameters), and \texttt{Bangla-LLaMA-2-7B}\footnote{\url{https://huggingface.co/BanglaLLM/bangla-llama-7b-instruct-v0.1}} (7B parameters). All models were used with their publicly available instruction-tuned checkpoints from Hugging Face and evaluated using their default decoding settings without additional fine-tuning.

\section{Performance Evaluation Metrics}\label{app:C}
\subsection{Keyword Extraction}
\noindent\textbf{MRR.} \textbf{M}ean \textbf{R}eciprocal \textbf{R}ank \citep{voorhees1999trec} assesses the average inverse position of the first accurately identified keyword within a group of queries. For each query, the reciprocal rank represents the multiplicative inverse of the position of the initial relevant keyword, and MRR is the mean of these values across all queries. A higher MRR indicates superior performance of the keyword extraction tool.\\

\noindent\textbf{mAP.} \textbf{M}ean \textbf{A}verage \textbf{P}recision \citep{baeza1999modern} evaluates the average precision of outcomes at every position in the ordered list. Precision denotes the proportion of pertinent results within the top returned outcomes. mAP is computed by averaging the precision scores for each position in the ordered list. Enhanced performance of the keyword extraction tool is indicated by a higher mAP.\\

\noindent\textbf{nDCG.} \textbf{N}ormalized \textbf{D}iscounted \textbf{C}umulative \textbf{G}ain \citep{jarvelin2017ir} evaluates the ranking quality of keyword extraction tools by taking into account the position of each relevant keyword, giving higher importance to keywords ranked higher on the list. It compares the relevance of the extracted keywords with an ideal list of keywords, and discounts the relevance of each keyword based on its position in the ranked list. The cumulative gain is then normalized against the ideal ranking to yield a score between 0 and 1, where 1 indicates the perfect ranking of keywords.

\subsection{Text Generation}
\noindent\textbf{BLEU.} \textbf{B}i\textbf{L}ingual \textbf{E}valuation \textbf{U}nderstudy \citep{papineni-etal-2002-bleu} assesses the similarity between reference and candidate texts by examining the overlap of their n-grams. It evaluates the number of matches between the generated and reference texts across different n-gram lengths (usually 1 to 4), and these matches are considered regardless of their position. The precision of these n-gram matches is calculated, and the values are aggregated using a geometric mean.\\

\noindent\textbf{ROUGE.} \textbf{R}ecall-\textbf{O}riented \textbf{U}nderstudy for \textbf{G}isting \textbf{E}valuation \citep{lin-2004-rouge} evaluates the quality of a generated text by comparing it to a reference text, assessing the overlap of content. ROUGE-N specifically gauges the overlap of N-grams between the generated and reference texts. For instance, ROUGE-1 evaluates the match of unigrams (individual words), while ROUGE-2 examines bi-gram overlap (pairs of words), and so forth. On the other hand, ROUGE-L assesses the longest common sub-sequence (LCS) between the generated and reference texts, capturing sentence-level structural similarities and automatically identifying the longest co-occurring in-sequence n-grams. In this study, we utilized ROUGE-1 and ROUGE-L metrics in terms of F-measure, which is the harmonic mean of precision and recall.\\

\noindent\textbf{BERTScore.} BERTScore \citep{zhang2019bertscore} employs the cosine similarity of contextual embeddings from a BERT-based model. The fundamental concept of BERTScore lies in contrasting the contextual embeddings of tokens within the generated text with those in the reference text, thereby considering the context of each word in the comparison. In this study, we used the \verb|bert-score|\footnote{\url{https://pypi.org/project/bert-score/}} library, which uses a multilingual BERT for Bangla texts.\\

\noindent\textbf{WER.} \textbf{W}ord \textbf{E}rror \textbf{R}ate \citep{errattahi2018automatic} evaluates the error rate by contrasting a generated text with a reference text known to be accurate. WER is computed using the Levenshtein distance, which tallies the number of single-character edits (insertions, deletions, or substitutions) needed to transform one word sequence into another. A lower WER denotes superior performance, with a perfect score of 0\% indicating exact correspondence between the generated and the reference text. Conversely, a higher WER signifies greater disparities between the generated and the reference text, indicating inferior performance.\\ 
\noindent\textbf{WIL.} \textbf{W}ord \textbf{I}nformation \textbf{L}ost \citep{errattahi2018automatic} is a metric similar to Word Error Rate (WER). While WER straightforwardly calculates errors based on insertions, deletions, and substitutions, WIL offers a distinct perspective by evaluating the amount of information lost or misrepresented during the text generation process. WIL is rooted in information theory and considers the probability of words as they occur in the language. The objective is to gauge errors by the informational content of the involved words, thereby assigning greater severity to errors on less predictable (more informative) words compared to more predictable ones. The logarithm of the probability, often represented as the base-2 logarithm, effectively quantifies the information content of the word. Consequently, words with higher predictability (and thus lower probability) convey less information. Consequently, WIL penalizes errors on less predictable words more significantly. Similar to WER, WIL is expressed as a percentage, with lower values indicating superior performance.

\section{Potential Applications of the Dataset}\label{app:D}
Having a dataset comprising $2.6$ million Bangla keyword-text pairs presents numerous opportunities for downstream tasks in Bangla Natural Language Processing (NLP) and Natural Language Generation (NLG). This dataset can be employed to evaluate keyword extraction tools, enabling the generation of concise summaries from large documents or text passages based on specified keywords. It will be interesting to extend the dataset to support other low-resource languages through multilingual machine translation techniques \citep{dabre2020survey}. Additionally, it can facilitate document classification tasks where the goal is to classify documents into predefined categories or topics based on their content (i.e. scientific articles into predefined categories). Keywords can serve as features for classification. Furthermore, the dataset can support information retrieval tasks by aiding in the retrieval of relevant documents or text passages based on user queries or keywords. Moreover, it can be utilized for training Named Entity Recognition (NER) models to identify and classify named entities mentioned in the text such as names of people, organizations, locations, dates, etc. With such a large dataset, question-answering and question-generation models can be trained to provide responses and generate questions accurately based on the provided keywords and contextual information from the text. Furthermore, text classification models can be trained to categorize text documents or passages into different classes or labels based on the provided keywords, enabling applications such as topic classification.

\section{Human Annotation}\label{app:E}
\subsection{Annotator Recruitment}
All the annotators involved in our study are native speakers of Bangla and possess experience in NLP. It is important to note that we assigned different annotators for keyword extraction (three) and evaluating generated texts (three). This means that annotators responsible for keyword extraction did not participate in evaluating text quality. For both tasks, we selected annotators based on their trustworthiness score \citep{price2020six}, which was determined through separate tests. Specifically, to recruit annotators for evaluating the quality of generated text, we provided them with $100$ keyword-text pairs and asked if the text fully contained the given keywords. From the test dataset, we randomly selected $80$ keywords-text pairs, while the remaining $20$ pairs served as control samples. The control samples consisted of \textit{keywords-generated text} pairs that were unknown to the participants beforehand. After completing the task, we analyzed the number of correctly labeled control samples for each annotator. A similar procedure was followed to recruit annotators for the keyword extraction task, where annotators were asked to extract important keywords from a text. Only annotators who achieved a trustworthiness score above $90\%$ were selected.

\subsection{Annotator Agreement}
\noindent\textbf{Kappa Score}. A statistical measure specifically designed for assessing agreement among multiple annotators. The Fleiss' kappa score \citep{fleiss1971measuring} is an extension of Cohen's kappa coefficient \citep{cohen1960coefficient} for situations where there are multiple annotators involved. The value of Fleiss' kappa can range from $-1$ to $1$. A value of $1$ signifies perfect agreement, while $0$ suggests that the annotator' judgments are independent of each other. Negative values indicate significant disagreement among the annotators.

\onecolumn

\section{Few Experimental Result Tables for Text Generation}\label{app:F}

% Table 1
\begin{table*}[h]
\centering
\resizebox{\textwidth}{!}{
\begin{tabularx}{\textwidth}{XX}

\hline
\textbf{Keys: [B]} {\bng অকাইঠারে, কৈয়া, কতা, নাই} & \textbf{{[}E]} idle, saying, words, no \\
\textbf{Original Text: [B]} {\bng অকাইঠারে বালা কতা কৈয়া লাব নাই।} & \textbf{{[}E]} There is no use in saying good words to the idle. \\
\textbf{mT5: [B]} {\bng অকাইঠারে কৈয়া কতা নাই।} & \textbf{{[}E]} There is nothing to say to the idle. \\
\textbf{BanglaT5: [B]} {\bng অকাইঠারে কতা কইয়া লাভ নাই।} & \textbf{{[}E]} There is no use in talking to the idle. \\
\textbf{Region:} Mymensingh & \\
\hline
\textbf{Keys: [B]} {\bng আমগো, নাম, পুরান, হুনছেন, ঢাকার} & \textbf{{[}E]} Our, name, old, hear, Dhaka \\
\textbf{Original Text: [B]} {\bng আমগো পুরান ঢাকার ছক্কু বেপারী লেনের নাম হুনছেন!} & \textbf{{[}E]} Have you heard the name of the Sakku Bepari lane of our old Dhaka? \\
\textbf{mT5: [B]} {\bng পুরান ঢাকার আমগো নাম হুনছেন ‘মৌসুমী’।} & \textbf{{[}E]} 'Mousumi' has heard the name of our old Dhaka. \\
\textbf{BanglaT5: [B]} {\bng আমগো পুরান ঢাকার নাম হুনছেন না?} & \textbf{{[}E]} Haven't you heard the name of our old Dhaka? \\
\textbf{Region:} Dhaka & \\
\hline
kl\textbf{Keys: [B]} {\bng কইলেই, লম্বর, নাম, পাইলে, দিবো, আমার, দেখায়} & \textbf{{[}E]} say, number, name, find, my, show \\
\textbf{Original Text: [B]} {\bng লম্বর খুইজা না পাইলে আমার নাম কইলেই মাইনছে দেখায় দিবো।} & \textbf{{[}E]} If you can't find the number, then people will show you when you say my name. \\
\textbf{mT5: [B]} {\bng লম্বর কইলেই আমার নাম দেখায়, পাইলে আমি দিবো না।} & \textbf{{[}E]} Saying the number shows my name; I won't give it if you find it. \\
\textbf{BanglaT5: [B]} {\bng আমার নাম কইলেই লম্বর দেখায়, আমি টাকা দিবো।} & \textbf{{[}E]} Saying my name shows the number, I will give the money. \\
\textbf{Region:} Dhaka & \\
\hline
\textbf{Keys: [B]} {\bng কড়ির, ভাগ, দেও, মোর} & \textbf{{[}E]} money, share, give, my \\
\textbf{Original Text: [B]} {\bng মোর পইসা কড়ির ভাগ মোক দেও।} & \textbf{{[}E]} Give me my share of the money. \\
\textbf{mT5: [B]} {\bng মোর কড়ির ভাগ দেও না।} & {[}E] Give share of my money. \\
\textbf{BanglaT5: [B]} {\bng মোর কড়ির ভাগ দেও, না হয়।} & {[}E] At least give me my share of the money. \\
\textbf{Region:} Barisal & \\
\hline
\end{tabularx}
}
\caption{The text generation capabilities of the fine-tuned \textit{mT5} and \textit{BanglaT5} models on dialectal data. Each row represent one data. \textbf{Region} denotes the geographical source of the dialectal data. \textbf{B} represents the keywords or texts in Bangla and \textbf{E} represents their corresponding English translation.}
\label{tab:dialect-data}
\end{table*}

% Table 2
\begin{table*}[t]
\centering
\resizebox{\textwidth}{!}{
\begin{tabularx}{\textwidth}{XX}

% \textbf{Cross Lingual keywords to text generation}                                                           & \textbf{English Translation}       \\
\hline
\textbf{Keys: [B]} {\bng patient, সপ্তাহ, বাড়ছে, হাসপাতালে, গত }                                                                & \textbf{{[}E]} patient, week, increasing, hospital, last                                                                                \\ 
\textbf{mT5: [B]} {\bng গত সপ্তাহ থেকে হাসপাতালে রোগীর সংখ্যা বাড়ছে।} & \textbf{{[}E]} The number of patients in the hospital has been increasing since last week.           \\ 
\textbf{BanglaT5: [B]} {\bng গত এক সপ্তাহ ধরে হাসপাতালে রোগীর সংখ্যা বাড়ছে।}                                                   & \textbf{{[}E]} The number of patients in the hospital has been increasing for the last one week.                                                               \\ \hline
\textbf{Keys: [B]} {\bng গ্রামের, teacher, একমাত্র}                                                              & \textbf{{[}E]} village, teacher, only                                                                             \\ 
\textbf{mT5: [B]} {\bng গ্রামের একমাত্র শিক্ষক।}                                                    & \textbf{{[}E]} The only teacher in the village.                                                     \\ 
\textbf{BanglaT5: [B]} {\bng গ্রামের একমাত্র শিক্ষক তিনি।}                                                      & \textbf{{[}E]} He is the only teacher in the village.                                                                    \\ \hline
\textbf{Keys: [B]} {\bng দেরিতে, মৃত্যু, doctor, রোগীর}                                                                & \textbf{{[}E]} delay, died, doctor, patient                                                                       \\
\textbf{mT5: [B]} {\bng দেরিতে থেকেই চিকিৎসাধীন রোগীর মৃত্যু হয়েছে বলে জানিয়েছেন হাসপাতালের একজন ডাক্তার।}     & \textbf{{[}E]} A doctor of the hospital said that the patient under treatment died due to delay.  \\ 
\textbf{BanglaT5: [B]} Doctor: {\bng রোগীর মৃত্যু দেরিতে হয়।}                                                        & \textbf{{[}E]} Doctor: Patient dies due to delay.                                                          \\ \hline
\textbf{Keys:} [B] {\bng প্রশ্নগুলোর, আজ, answer}                                                                  & \textbf{{[}E]} questions, today, answer                                                                                  \\ 
\textbf{mT5:} [B] {\bng আজ আপনার প্রশ্নগুলোর উত্তর পাবেন।}                                                               & \textbf{{[}E]} You will get the answers to your questions today.                                                                           \\ 
\textbf{BanglaT5:} [B] {\bng আজ প্রশ্নগুলোর Answer দেবেন।}                                           & \textbf{{[}E]} Answer the questions today.    \\  \hline
\textbf{Keys:}  village, teacher, only                                                                  &                                                                                   \\ 
\textbf{mT5:} [B] {\bng আমিও একজন স্কুলশিক্ষক, শিক্ষক-শিক্ষার্থীরই (বিদ্যালয়)।}                                                               & \textbf{{[}E]} I am also a school teacher, (school) is for teachers and students.                                                                           \\ 
\textbf{BanglaT5:} [B] {\bng শুধু একটা স্কুলশিক্ষক গ্রামে।}                                           & \textbf{{[}E]} There is only a school teacher in the village.    \\  \hline
\textbf{Keys:} increase, hospital, last                                                               &   \\ 
\textbf{mT5:} [B] {\bng এর আগে হাসপাতালেই মৃত্যুর সংখ্যা increase।}                                                               & \textbf{{[}E]} Before this, the number of deaths in the hospital increased.                                                                         \\ 
\textbf{BanglaT5:} [B] {\bng সর্বশেষ গত বছর hospitalize সূচক বেড়েছে।}                                           & \textbf{{[}E]}The hospitalize index increased last year.    \\  \hline
\textbf{Keys:} questions, today, answer                                                                 &    \\ 
\textbf{mT5:} [B] {\bng কিন্তু আপনার প্রশ্নগুলোও আজকালই ‘উত্তর’।}                                                               & \textbf{{[}E]} But your questions are also 'answers' nowadays.                                                                           \\ 
\textbf{BanglaT5:} [B] {\bng আজকে প্রশ্নের উত্তর দাও।}                                           & \textbf{{[}E]} Give answers to the questions today.    \\  \hline
\textbf{Keys:} come, police                                                              &                                                                             \\ 
\textbf{mT5:} [B] {\bng পুলিশও আছে।}                                                               & \textbf{{[}E]} There are also police.                                                                           \\ 
\textbf{BanglaT5:} [B] {\bng police এসে যাবে।}                                           & \textbf{{[}E]} The police will come.    \\  \hline
\end{tabularx}
}
\caption{Few instances demonstrating the cross-lingual transfer ability in the generated texts produced by both the fine-tuned models (\textit{mT5} and \textit{BanglaT5}) when provided with cross-lingual keywords. \textbf{B} represents the keywords or texts in Bangla and \textbf{E} represents their corresponding English translation.}
\label{tab:cross-lingual-transfer}
\end{table*}

% Table 3
\begin{table*}[t]
\centering
% \resizebox{\columnwidth}{!}{
\begin{tabular}{ccc} 
\hline
\textbf{Keyword}                                                  & \textbf{mT5 Text}                                                                       & \textbf{BanglaT5 Text}                                                                   \\ 
\hline
\begin{tabular}[c]{@{}c@{}}\bng{গতকাল} \\(yesterday)\end{tabular} & \begin{tabular}[c]{@{}c@{}}\bng{গতকাল শুক্রবার।} \\(Yesterday was Friday.)\end{tabular} & \begin{tabular}[c]{@{}c@{}}\bng{গতকাল শনিবার।} \\(Yesterday was Saturday.)\end{tabular}  \\ 
\hline
\begin{tabular}[c]{@{}c@{}}\bng{আসবে} \\(will come)\end{tabular}  & \begin{tabular}[c]{@{}c@{}}\bng{আসবে কীভাবে?} \\(How come?)\end{tabular}                & \begin{tabular}[c]{@{}c@{}}\bng{পুলিশ আসবে।} \\(The police will come.)\end{tabular}      \\ 
\hline
\begin{tabular}[c]{@{}c@{}}\bng{এবার} \\(now)\end{tabular}        & \begin{tabular}[c]{@{}c@{}}\bng{এবার কী?} \\(What now?)\end{tabular}                    & \begin{tabular}[c]{@{}c@{}}\bng{এবার ব্যতিক্রম।} \\(Now is an exception.)\end{tabular}   \\ 
\hline
\begin{tabular}[c]{@{}c@{}}\bng{নববর্ষ} \\(New Year)\end{tabular} & \begin{tabular}[c]{@{}c@{}}\bng{নববর্ষ শুরু।} \\(New Year begins.)\end{tabular}         & \begin{tabular}[c]{@{}c@{}}\bng{শুভ নববর্ষ।} \\(Happy New Year.)\end{tabular}            \\
\hline
\end{tabular}
\caption{Few examples of generated texts by both fine-tuned models with a single keyword.}
\label{tab:single-key}
\end{table*}

% Table 4
\begin{table*}[t]
\centering
\resizebox{\textwidth}{!}{
\begin{tabularx}{\textwidth}{XX}

\hline
\textbf{Keys: [B]} {\bng সঙ্গে, হয়তো, সন্ধ্যার, স্বাভাবিকভাবে, মেয়র, বহু, বিষয়ে, বলেন, মধ্যে, পাওয়া, আজ, জবাব }                                                                & \textbf{{[}E]} with, may be, evening, naturally, Mayor, many, regard, said, by, available, today, answer \\ 
\textbf{mT5: [B]} {\bng আজ সন্ধ্যার মধ্যে হয়তো স্বাভাবিকভাবে জবাব পাওয়া যাবে। এ বিষয়ে আরও বলেন, মেয়র মোশাররফ হোসেনের সঙ্গে বহু দিন ধরে আলোচনা চলছে।} & \textbf{{[}E]} The answer may be available naturally by evening today. In this regard, he further said that discussions are going on with Mayor Mosharraf Hossain for many days.           \\ 
\textbf{BanglaT5: [B]} {\bng আজ সন্ধ্যায় মেয়র প্রার্থীদের সঙ্গে আলোচনার বিষয়টি নিশ্চিত করে আরও বলেন, তাঁদের স্বাভাবিক প্রতিক্রিয়া হয়তো অনেক ক্ষেত্রেই পাওয়া যায়। এ বিষয়ে মেয়রের ভাষ্য ছিল অনেকেরই কোনো জবাব নেই!}                                                   & \textbf{{[}E]} Confirming the discussion with the Mayoral candidates today evening, they said their natural response may be available in many cases. The mayor's comment in this regard was that many people have no answer!\\ \hline
\textbf{Keys: [B]} {\bng মোটরসাইকেলে, শিক্ষক, ছাত্রীর, থাকা, এক, চালকের, মোড়ে, কয়েকজন, গিয়ে, ও, বাসে, একটি, তাঁদের}                                                              & \textbf{{[}E]} on the motorcycle, teacher, of the student, staying, one, of the driver, at the corner, a few, went, and, on the bus, one, theirs     \\ 
\textbf{mT5: [B]} {\bng এরপর কয়েকজন শিক্ষক মোড়ে গিয়ে তাঁদের বাসে ও একটি মোটরসাইকেলে আগুন দেন। পুলিশ জানায়, ঘটনাস্থলে থাকা এক ছাত্রীর চালকের সঙ্গে কথা-কাটাকাটি হয়। }                                                    & \textbf{{[}E]} Afterwards, a few teachers went to the corner on the bus and the motorcycle, and set fire. Police informed, some conversation happened between a student of the driver and the teachers.                                                 \\ 
\textbf{BanglaT5: [B]} {\bng তাঁদের মধ্যে কয়েকজন শিক্ষক ও চালকের সহযোগিতায় একটি বাসে গিয়ে মোটরসাইকেলে থাকা এক ছাত্রীর লাশ উদ্ধার করে ধানমন্ডি মোড়ে পাঠানো হয়।} & \textbf{{[}E]} Among them, with the cooperation of a few teachers and the driver, a body of a student staying on the motorcycle and the bus was retrieved from Dhanmondi corner and sent.  \\ \hline
\end{tabularx}
}
\caption{Few instances demonstrating the both fine-tuned models (\textit{mT5} and \textit{BanglaT5}) generate texts spanning across multiple sentences. \textbf{B} represents the keywords or texts in Bangla and \textbf{E} represents their corresponding English translation.}
\label{tab:single-line-test}
\end{table*}

% \section{Language Resource References}
% \label{lr:ref}
% \bibliographystylelanguageresource{lrec2026-natbib}
% \bibliographylanguageresource{languageresource}

\end{document}